\documentclass[10pt,twocolumn,letterpaper]{article}

\usepackage{cvpr}              

\usepackage[export]{adjustbox}
\usepackage{array}
\usepackage{multirow}
\usepackage{arydshln}
\usepackage{amssymb}
\usepackage{MnSymbol}
\usepackage{pifont}
\newcommand{\cmark}{\ding{51}}%
\newcommand{\xmark}{\ding{55}}%
\usepackage{booktabs}
\usepackage[dvipsnames]{xcolor}
\usepackage[nolist]{acronym}
\usepackage{lineno} 
\usepackage{csquotes}
\usepackage{tikz}
\usepackage{siunitx}
\usepackage[inkscapelatex=false]{svg}
\usepackage{mathtools}

\begin{acronym}
	\acro{HD}{high-definition}
\end{acronym}

\DeclareMathAlphabet\mathbfcal{OMS}{cmsy}{b}{n}

\DeclarePairedDelimiter\norm{\lVert}{\rVert}

\definecolor{cvprblue}{rgb}{0.21,0.49,0.74}
\usepackage[pagebackref,breaklinks,colorlinks,allcolors=cvprblue]{hyperref}

\def\paperID{9} 
\def\confName{CVPR}
\def\confYear{2026}

\title{XD-MAP: Cross-Modal Domain Adaptation via Semantic Parametric\\Maps for Scalable Training Data Generation}
\author{
Frank Bieder$^{1,2}$ \quad Hendrik Königshof$^{2}$ \quad Haohao Hu$^{1}$ \quad 
Fabian Immel$^{1,2}$  \\ Yinzhe Shen$^{1}$ \quad Jan-Hendrik Pauls$^{1}$ \quad  Christoph Stiller$^{1}$ \\
$^{1}$KIT Karlsruhe Institute of Technology \quad $^{2}$FZI Research Center for Information Technology\\
{\tt\small \{firstname.lastname\}@kit.edu} \quad {\tt\small \{lastname\}@fzi.de}
}

\begin{document}
\maketitle
\begin{abstract}

Until open-world foundation models match the performance of specialized approaches, deep learning systems remain dependent on task- and sensor-specific data availability.
To bridge the gap between available datasets and deployment domains, domain adaptation strategies are widely used.
In this work, we propose XD-MAP, a novel approach to transfer sensor-specific knowledge from an image dataset to LiDAR, an entirely different sensing domain.
Our method leverages detections on camera images to create a semantic parametric map.
The map elements are modeled to produce pseudo labels in the target domain 
without any manual annotation effort.
Unlike previous domain transfer approaches, our method does not require direct overlap between sensors and enables extending the angular perception range from a front-view camera to a full \SI{360}{\degree} view.
On our large-scale road feature dataset, XD-MAP outperforms single shot baseline approaches by +19.5 mIoU for 2D semantic segmentation, +19.5 $\textrm{PQ}_\textrm{th}$ for 2D panoptic segmentation, and +32.3 mIoU in 3D semantic segmentation. The results demonstrate the effectiveness of our approach achieving strong performance on LiDAR data without any manual labeling.
\end{abstract}    
\section{Introduction}
\label{sec:introduction}

Deep learning models for perception rely heavily on large-scale annotated datasets, which are often constrained to specific sensing modalities or even sensor models. 
This limitation poses a challenge for deploying models across diverse sensor configurations or changing sensor setups. 
Many existing domain adaptation approaches fall into two categories: (i) Techniques that align the source and target domains by adapting sensor-specific characteristics, such as the ray distribution in LiDAR-to-LiDAR adaptation, and (ii) methods that exploit overlapping sensor fields of view to transfer knowledge from simultaneously perceived objects.
While effective under specific conditions, both approaches impose strong requirements, either on the similarity of sensor modalities or the existence and extent of shared fields of view, limiting their applicability to real world sensor setups.

To overcome these limitations, we propose XD-MAP, a self-supervised cross-modal domain adaptation framework that enables knowledge transfer using a semantic high-definition (HD) map.
Based on a feature point SLAM, we construct a precise map consisting of semantic landmarks, such as poles, traffic signs, and traffic lights.
One key point is to use a suitable parametric geometric representation for each semantic class, specifically planes and cylinders.
By projecting these semantic primitives into the target sensor domain, we generate dense, structured ground-truth labels for the target modality.
Our approach supports multiple label representations, including point-wise 3D segmentation by assigning class labels to individual LiDAR points and range image-based annotation, preserving the unique rendered 2D shape of objects within the ground truth data.

\begin{figure*}
\centering
   \includegraphics[trim={0 0.5cm 0 0.1cm},clip, width=\linewidth]{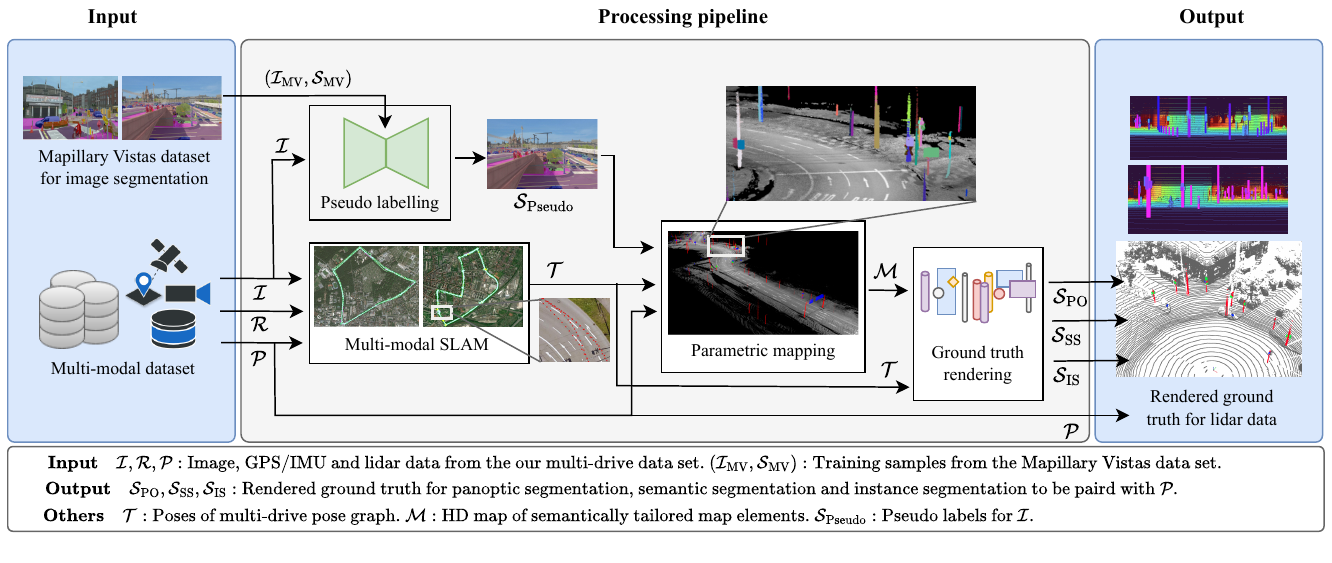}
    \caption{
    Overview of the proposed XD-MAP:
    We generate pseudo labels from RGB camera images from a single camera using a neural network trained on Mapillary Vistas~\cite{Neuhold_Mapillary_2017_ICCV}, a well-generalizing image dataset, as source.
    Using an accurate SLAM and parametric geometric representations, tailored to the semantic classes of objects, we build a precise semantic map of objects of interest.
    The mapped objects can then be rendered into the target domain, in our case a LiDAR sensor, opening up a new sensing modality and extending the perception range from a front-view camera to \SI{360}{\degree} coverage.
    As exemplary tasks, we cover 2D panoptic as well as 2D and 3D semantic segmentation.
    }
    \label{fig:overview}
\end{figure*}

The main contribution of XD-MAP is a self-supervised domain adaptation pipeline that requires neither overlapping sensor fields of view nor modal similarities.
A semantic parametric map serves as bridge between detections of a pretrained neural network as source and the target domain.
This makes it independent from direct sensor overlap and suitable for \SI{360}{\degree} LiDAR perception. 
We show that compared to single shot baselines, our approach strongly increases the performance of segmentation models and produces accurate detections. 
In ablation studies, we investigate the influence of map element range and sensor frequency on performance, both important for practical considerations in data collection or downstream tasks.
The proposed XD-MAP framework thus facilitates a scalable and adaptable pipeline for multi-sensor autonomous perception, bridging the gap between heterogeneous sensor modalities, transferring knowledge to sensors without suitable datasets, and extending the perceptive field of view.
An overview is depicted in \cref{fig:overview}.

%

\section{Related Work}
\label{sec:related_work}


\paragraph{Semantic Parametric Maps}
\label{sec:semantic_primitive_mapping}

Due to their relevance for automated driving, road signs have been detected and mapped early on~\cite{Soheilian2013,Vu2013}, already using planes in 3D space as representation. Traffic lights are commonly detected in camera images since this allows estimating their lights' state~\cite{Fairfield2011,Levinson2011}.
Usually, they are mapped as a point with an optional height or a number of lights. Purely geometric primitives, \eg edges~\cite{Zhang2014}, flat surfaces~\cite{Zhang2014,GeometricPrimitives_Kuemmerle_2019}, and poles~\cite{PoleLike_Sefati_2017,GeometricPrimitives_Kuemmerle_2019}, have been proposed as compact yet robust semantic features for SLAM and localization and are often extracted from stereo cameras or LiDAR point clouds.
They are represented as planes or lines, occasionally with a diameter for poles.

In one of the first generic approaches, \cite{ParametricMapping_Pauls_2021} proposed to detect poles, traffic lights and road signs in a camera image using a neural network.
Then, a parametric representation can be estimated accurately by fusing the detection with LiDAR measurements.
While deviating in the landmark representation and map optimization, we adapt the idea to branch the parametric representation based on the semantic class of an object.

\paragraph{Semantic and Panoptic Segmentation} 

Semantic segmentation and panoptic segmentation are well-studied computer vision tasks.
Various surveys provide a good overview of the current state of the art of semantic segmentation methods, for example categorized according to their supervision level~\cite{hao2020brief} or summarizing all approaches using vision transformers~\cite{thisanke2023semantic}.
There are also numerous surveys in the field of object detection which show the development of work in recent years from CNNs to transformers~\cite{zou2023object,arkin2023survey}.
Recent work such as Mask2Former~\cite{mask2former} and Oneformer~\cite{jain2023oneformer} remove the need for specialized architectures and unify both tasks on an architecture or model level.

Inspired by the success of transformers~\cite{vaswani2017attention} in the field of computer vision, several researchers have attempted to apply this architecture to point clouds.
Due to the sparse nature of point clouds, various sparse attention strategies have been explored.
VoTr~\cite{mao2021voxel} introduces local and dilated attention using a custom voxel query to enable attention mechanisms on sparse voxels.
Cylinder3D~\cite{zhu2021cylindrical} uses cylindrical partition for voxels and asymmetrical 3D convolution to better fit the general and object point distributions.
SST~\cite{fan2022embracing} divides the 3D space into regional groups, so that the self-attentions only interact with tokens coming from the same regions.
DSVT~\cite{wang2023dsvt} efficiently processes point clouds using a dynamic sparse window attention mechanism and a rotation-based partitioning strategy.

\begin{figure*}
    \centering
    \begin{subfigure}{0.497\textwidth}
        \includegraphics[trim={0 6cm 0 1cm},clip, width=\columnwidth]{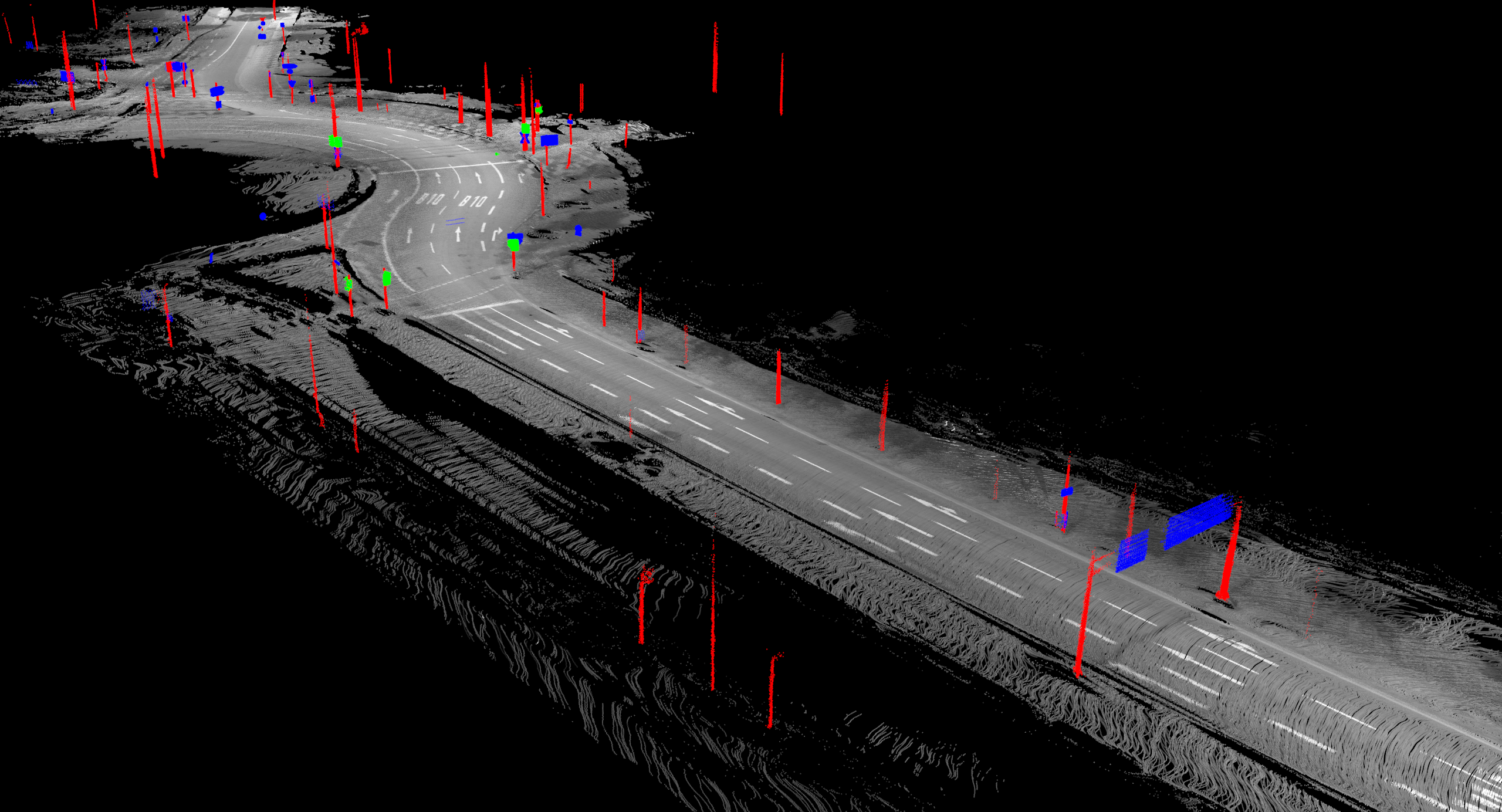}
    \end{subfigure}
    \hfill
    \begin{subfigure}{0.497\textwidth}
        \includegraphics[trim={0 6cm 0 1cm},clip, width=\columnwidth]{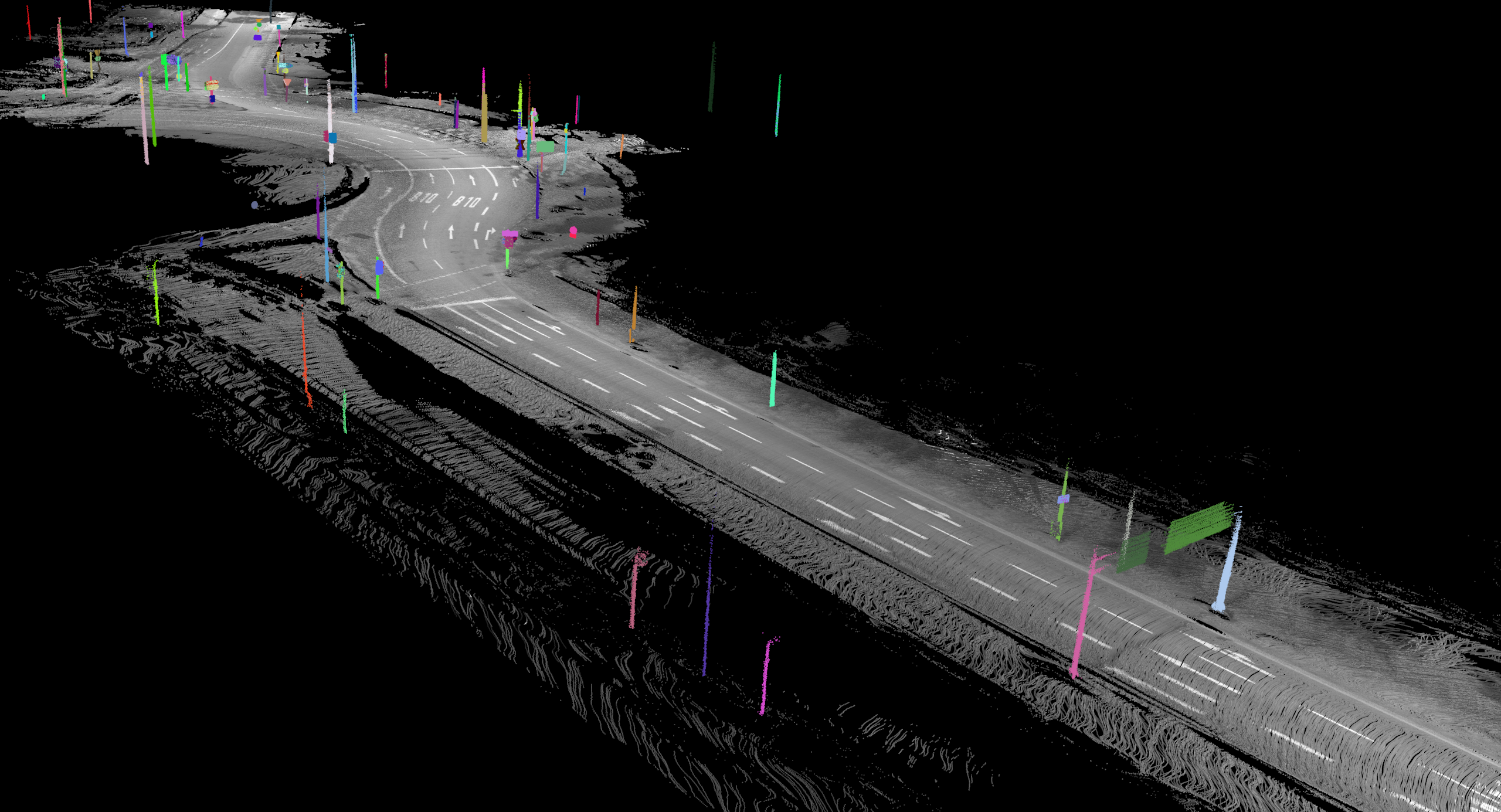}
    \end{subfigure}
    \hfill
    \begin{subfigure}{0.33\textwidth}
        \includegraphics[trim={0 0 0 5cm},clip, width=\columnwidth]{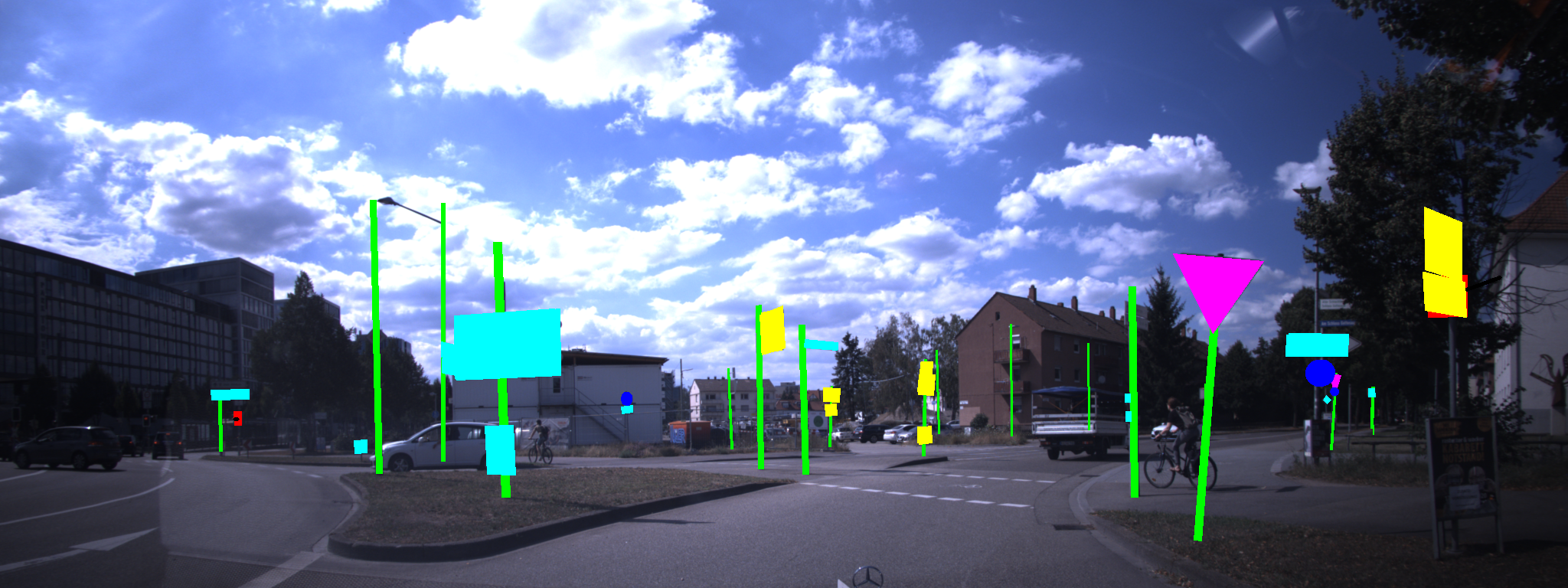}
    \end{subfigure}
    \begin{subfigure}{0.33\textwidth}
        \includegraphics[trim={0 0 0 5cm},clip, width=\columnwidth]{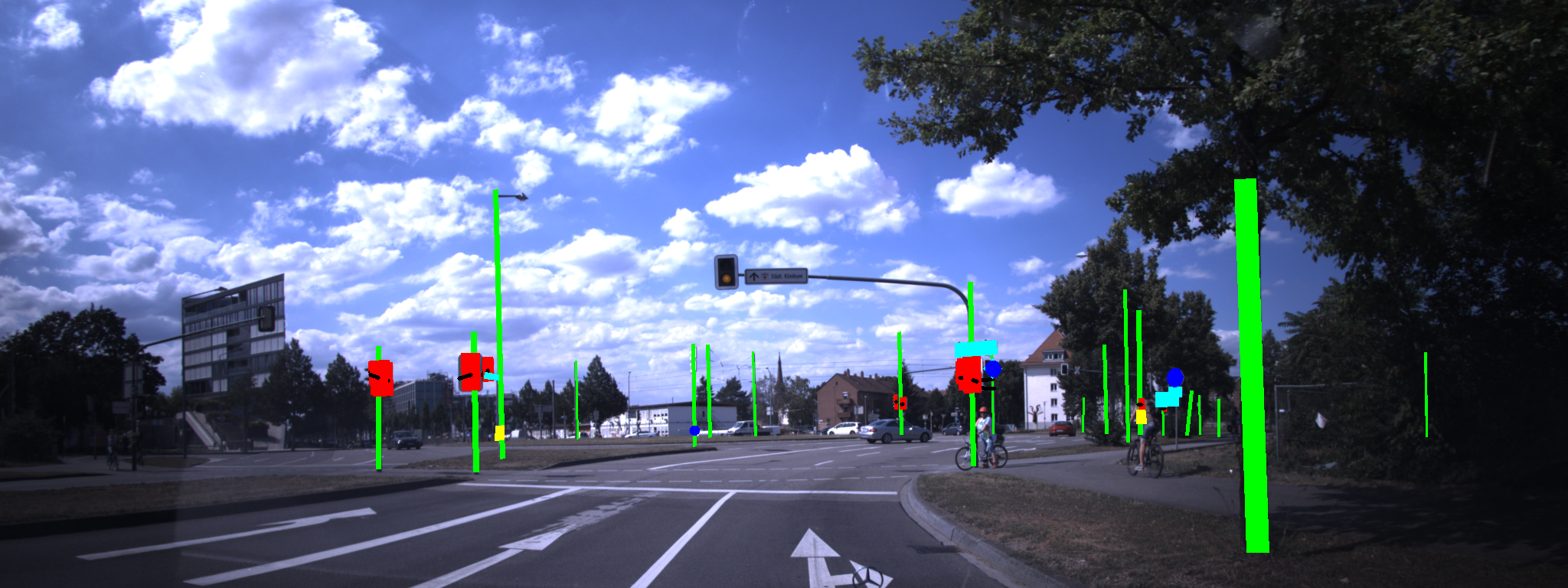}
    \end{subfigure}
    \hfill
    \begin{subfigure}{0.33\textwidth}
        \includegraphics[trim={0 0 0 5cm},clip, width=\columnwidth]{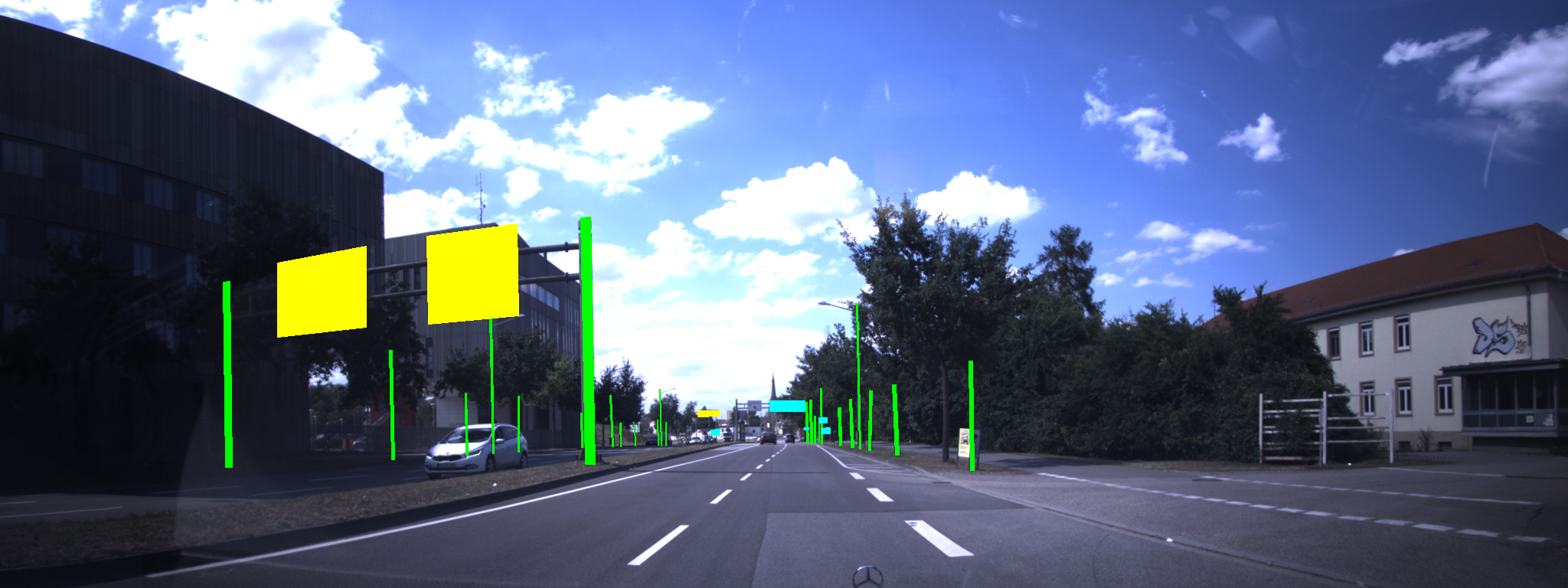}
    \end{subfigure}
    \caption{
        Exemplary results of the semantic parametric mapping.
        Depicted are parametric primitives of three semantic classes (poles, traffic lights, road signs).
        The parametric map provides pseudo-labels for an accumulated LiDAR point cloud colored by semantic class and instance, respectively (top row).
        As sanity check or to transfer between cameras, the map elements can also be projected into camera images (bottom row).
        Best viewed with digital zoom.
    }
    \label{fig:parametric_mapping}
\end{figure*}

\paragraph{Cross-Sensor Domain Adaptation}
Cross-sensor domain adaptation has gained significant attention in various applications, particularly in autonomous driving and remote sensing.
SOAP~\cite{huang2024soap}  improves LiDAR-based 3D object detection by generating high-quality pseudo-labels for stationary objects, achieving notable domain gap reduction.
CroMA~\cite{mancroma}  focuses on monocular BEV perception, using a LiDAR-teacher and camera-student model to transfer knowledge across modalities.
Similarly, SRDAN~\cite{zhang2021srdan} leverages geometric characteristics to align features in 3D object detection across datasets.
For satellite imagery, methods like CyCADA~\cite{mateo2020cross} and LoveCS~\cite{wang2022cross} address cross-sensor issues by transforming data to reduce statistical differences, enhancing transfer learning models for remote sensing applications.
Techniques such as FDA~\cite{prabhakar2022} and the supervised pre-training approach in~\cite{rist2019} also contribute to improved cross-domain classification and feature transferability.
In the context of semantic segmentation, using geometric mapping systems improves generalization across LiDAR sensors~\cite{langer2020domain}.
An unsupervised domain adaptation technique for multi-modal LiDAR segmentation, employing depth completion to align 2D image features across domains, was proposed in \cite{cardace2023boosting}.
A comprehensive review of recent progress in domain adaptation methods targeted towards LiDAR perception is presented in \cite{triess2021survey}.

\section{Semantic Parametric Mapping}
\label{sec:para_mapping}
In order to bridge the gap between the sensor domains, we exploit the fact that certain elements in the environment are static. 
Hence, we estimate a map consisting of geometric primitives that represent static objects from three exemplary semantic classes.
Coarsely inspired by~\cite{ParametricMapping_Pauls_2021}, the shape of geometric primitives depends on the respective semantic class as detected by a neural network:
Poles and traffic lights are modeled as cylinders while road signs are upright planes of various shapes.

The elements are detected using a well-generalizing neural network~\cite{SeamSeg_Porzi_2019}, yielding an instance mask in the camera image for each object.
Using it, a map element-specific point cloud can be determined by projecting all points into the masked camera image.
Assuming an accurate ego motion determined using a feature-based SLAM system~\cite{Sons2017,Sons2018}, detections can be associated in the global map frame.

In contrast to \cite{ParametricMapping_Pauls_2021}, we exploit all instance masks and element point clouds to estimate a single map element.
Given the shape of the parametric model, \eg a cylinder for a pole, the model parameters can be optimized by minimizing deviations between a) the map element-specific point clouds and b) viewing rays on the instance mask's contour pixels and the object shape's hull using a non-linear optimizer~\cite{Agarwal_Ceres_Solver_2022}.
For traffic signs with known shapes, \ie rectangles, circles, and triangles, a shape prior is induced.

This results in a highly accurate and complete map.
Using the known ego pose in the map, the 3D shapes can be projected back into the original camera images, yielding pixel-accurate instance masks as depicted in \cref{fig:parametric_mapping}.
More interestingly, they can be used to weakly annotate a LiDAR point cloud for other tasks, as presented in the next section.

\section{Pseudo Label Generation}

We present our proposed approach, XD-MAP, as well as two baselines that use single shot pseudo labels instead of a semantic parametric HD map.

\subsection{XD-MAP}

\paragraph{Spherical Camera Model} 
Following \cite{Milioto2029}, we transform point clouds into a spherical projection. Our model extends the standard spherical projection by incorporating independent horizontal and vertical resolutions, aligned with the respective minimum angular resolutions of the LiDAR sensor.

\paragraph{Pseudo Label Generation}
For each ego pose $T_i$, we obtain a set of landmarks \( \mathcal{L}_{i} = \{ l \in \mathcal{M} \mid \norm{l - T_i}_2 < \tau \} \) from our semantic map \(\mathcal{M}\), where the distance between a landmark \(l\) and the ego pose is below a predefined threshold \(\tau\).
Key points of each landmark are projected using the spherical camera model, and the landmark’s shape is reconstructed in sensor space, producing a polygon set \(S\) representing object boundaries.
A map element is retained for further processing if at least one polygon point falls within the sensor’s region of interest.
Segments smaller than a critical pixel threshold are filtered out.

To account for occlusions, the map elements are processed in decreasing distance from the sensor.
Each newly projected element overwrites previously projected ones, ensuring that objects are correctly represented from the sensor's perspective.
From this representation, pseudo labels for various 2D perception tasks, such as object detection, semantic segmentation, and panoptic segmentation, can be derived directly.

Our 3D point cloud annotation utilizes the same intermediate annotation and combines each 2D shape with an object-specific depth, creating a frustum that encapsulates the corresponding map element.
All LiDAR points falling within this frustum are assigned the label of the map element.
This method ensures that the 3D labels accurately reflect the semantic information from the map while maintaining spatial consistency within the point cloud.

\begin{figure}
    \centering
    \begin{subfigure}{0.23\textwidth}
        \includegraphics[width=\columnwidth]{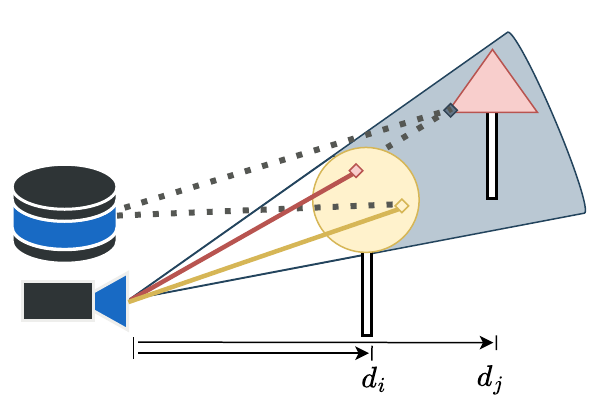}
        \subcaption{
            Parallax effect due to different optical centers of LiDAR and camera.
        }
        \label{fig:lidar_processing:p}
    \end{subfigure}
    \hfill
    \begin{subfigure}{0.23\textwidth}
        \includegraphics[width=\columnwidth]{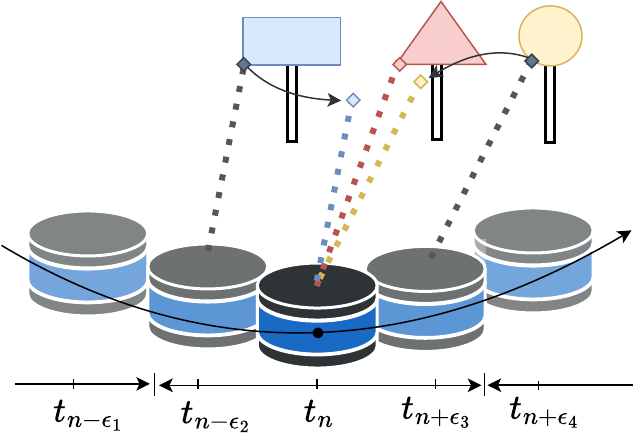}
        \subcaption{
            Rolling shutter artifacts require motion compensation.
        }
        \label{fig:lidar_processing:mc}
    \end{subfigure}
    \caption{
        Illustration of measurement artifacts affecting our pipeline.
        The parallax effect leads to background LiDAR points appearing on foreground objects when projected in the camera space. 
    }
    \label{fig:lidar_processing}
\end{figure}

\subsection{Baseline Approaches}
We implement two single-shot baselines, XD-B1 and XD-B2. The first approach specializes in preserving the object's shape, while the second aims to retain the precise LiDAR depth for each point within each object. The key challenge these approaches address is how to accurately infer both the 3D position and spatial extent of an object in LiDAR space when it is initially represented as a 2D polygon in the camera image. Given that the camera image provides a limited field of view, the baselines can only provide annotations for roughly \SI{100}{\degree}.  Within this region, object instances detected in the camera image are projected into LiDAR space and annotated accordingly, while non-object points are labeled as background to clearly distinguish them. Additionally, points or pixels outside the supervised FOV are marked to be ignored during training to avoid negatively affecting the optimization process. 

\paragraph{XD-B1: Shape-Preserving Lifting}
The first baseline lifts 2D front-view instance masks into 3D by assigning a single representative LiDAR depth to each instance. This depth is estimated using the 30th percentile of the depths of the LiDAR points falling within the instance. Selecting the 30th percentile instead of the mean mitigates parallax effects (\cref{fig:lidar_processing:p}), as background LiDAR points may be incorrectly projected onto foreground objects in camera space. The lifted instance polygons are then reprojected into the spherical LiDAR representation, preserving the geometric contours of the objects.

\paragraph{XD-B2: Depth-Preserving Projection}
In this baseline, the 2D semantic instance masks function as a lookup table for associating instance labels to LiDAR points. 
Instead of estimating a single depth, all LiDAR points corresponding to an instance are used to construct a convex hull, which serves as a pseudo labeling mechanism. 
This prioritizes the accurate retention of depth information for each LiDAR point within the instance, but does not enforce any shape.

\subsection{Uncertainty Mitigation}
\label{subsec:uncertainty}
In the annotation process, small errors in localization or calibration can cause points near contours to fall outside a primitive.
Since objects such as traffic lights, traffic signs, and poles are sparsely distributed in the scene, we prioritize reducing false negative instance annotations, even at the cost of introducing false positives.
To address this, in the annotation process of 3D point clouds, we introduce an uncertainty margin by expanding the radius of each cylindrical object by \SI{5}{\centi\meter} and \SI{7}{\centi\meter} for traffic lights and poles, respectively.
Planes are extruded to a 3D box by assuming a width of \SI{10}{\centi\meter}.

For 2D labels, we apply a minimal dilation operation to the rendered shape, resulting in a 1-pixel expansion of each segment.
This adjustment also prevents artifacts such as poles from being rendered with a width of zero.









\section{Experimental Evaluation}
\label{sec:evaluation}

\subsection{Dataset, Task Definition and Metrics}

\begin{figure}
    \includegraphics[width=\columnwidth]{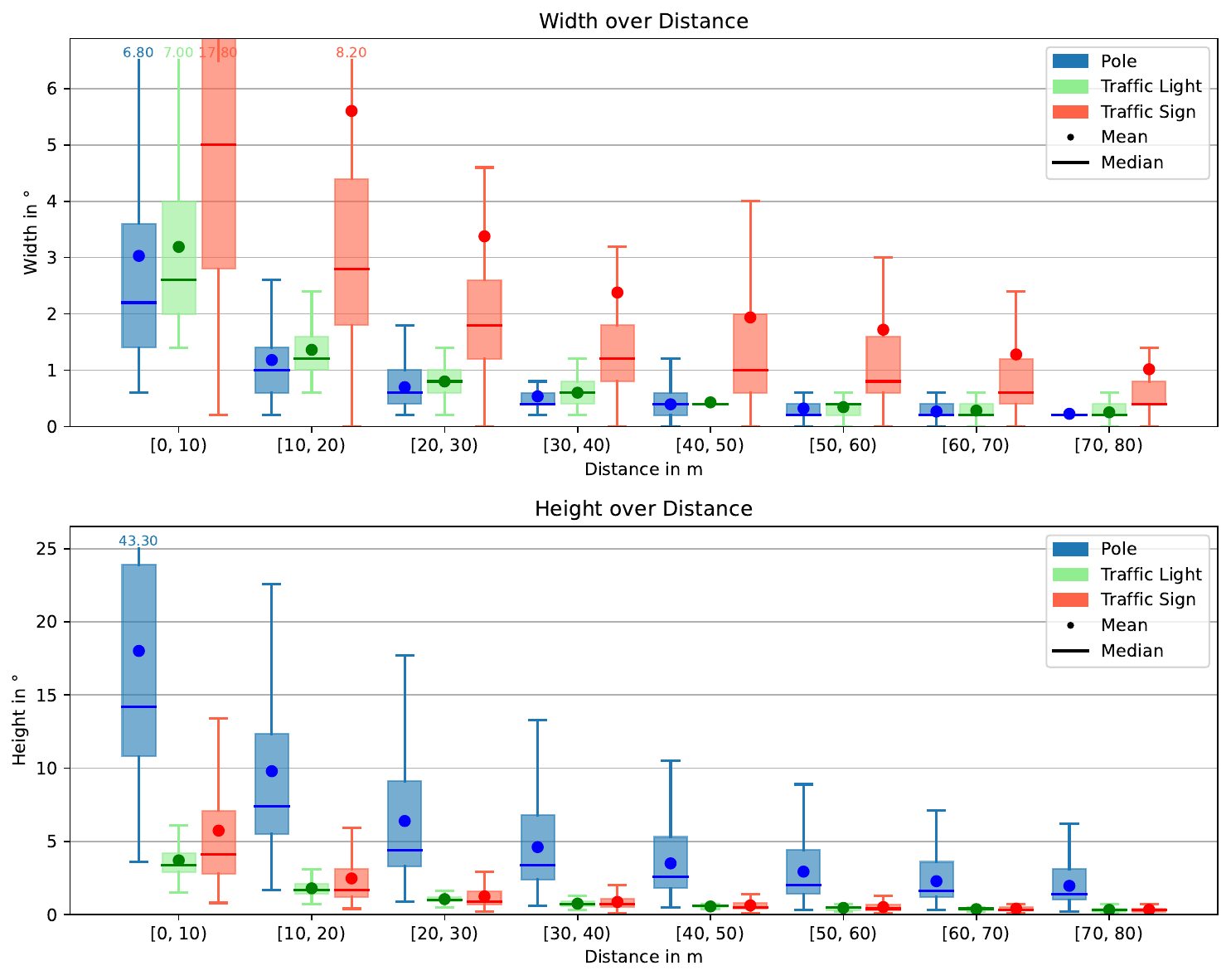}
    \caption{Boxplot of instance height and width in the field of view of the LiDAR as a function of its distance to the ego vehicle. }
    \label{fig:boxplot}
\end{figure}

\paragraph{Sensor Characteristics} 
A high sensor resolution and a precise calibration are essential for the proposed pipeline, particularly given the specific selection of semantic objects. When back-projecting a 10 cm-wide object from a distance of 50 m, even minor errors can significantly impact accuracy. This imposes strict resolution requirements that many publicly available datasets fail to meet. Typical object extensions in our dataset can be reviewed in \cref{fig:boxplot}.
Hence, our sensor suite includes a Velodyne Alpha Prime LiDAR with 128$\times$1812 rays per scan and an RGB camera with a resolution of 1536$\times$4096 pixels.
The sensors are synchronized via an external trigger operating at 10 Hz and accurately calibrated using checkerboards and a spherical target~\cite{kummerle_automatic_2018,beck_generalized_2018,straus_calibrating_2014}.
%
%



\begin{table}[t]
\centering
\caption{Summary of all sequences regarding total kilometers, number of frames of each drive (f) in train and test split and number of mapped instances (n).}
\label{tab:DataRecording}
\resizebox{0.95\columnwidth}{!}{
\begin{tabular}
{@{\extracolsep{\fill}}l cccc cccc@{}}
\toprule
\multirow{2}{*}{\textbf{\#}}  && \multicolumn{3}{c}{\textbf{Data Recording}} && \multicolumn{3}{c}{\textbf{Parametric Map}}  \\ 
\cmidrule{3-5}
\cmidrule{7-9}
&&{$\mathbf{Length}$} &{$\mathbf{f_{train}}$} &{$\mathbf{f_{eval}}$} &&{$\mathbf{n_{TS}}$} & {$\mathbf{n_{TL}}$} & {$\mathbf{n_{P}}$} \\ 
\midrule
$\mathrm{Seq}_1$ &&5.35 km &3973 & 1336 & &897& 367 & 640    \\ 
$\mathrm{Seq}_2$ &&2.97 km & 3913 & 0 && 479 &170 & 369     \\ 
$\mathrm{Seq}_3$ &&2.98 km & 4085& 0 && 449& 163& 323    \\ 
$\mathrm{Seq}_4$ &&5.33 km & 4778& 0 &&800 & 159& 547   \\ 
$\mathrm{Seq}_5$ &&4.95 km & 3722& 972 &&829 &265 & 474 \\ 
\midrule
\textbf{Total}  &&21.58 km & 20471& 2308& & 2353 & 1124 & 3454  \\
\bottomrule
\end{tabular}
}
\label{table_map}
\end{table}
\begin{figure}
\definecolor{darkyellow}{RGB}{255,220,0}
\centering
    \centering
    \includegraphics[width=\columnwidth]{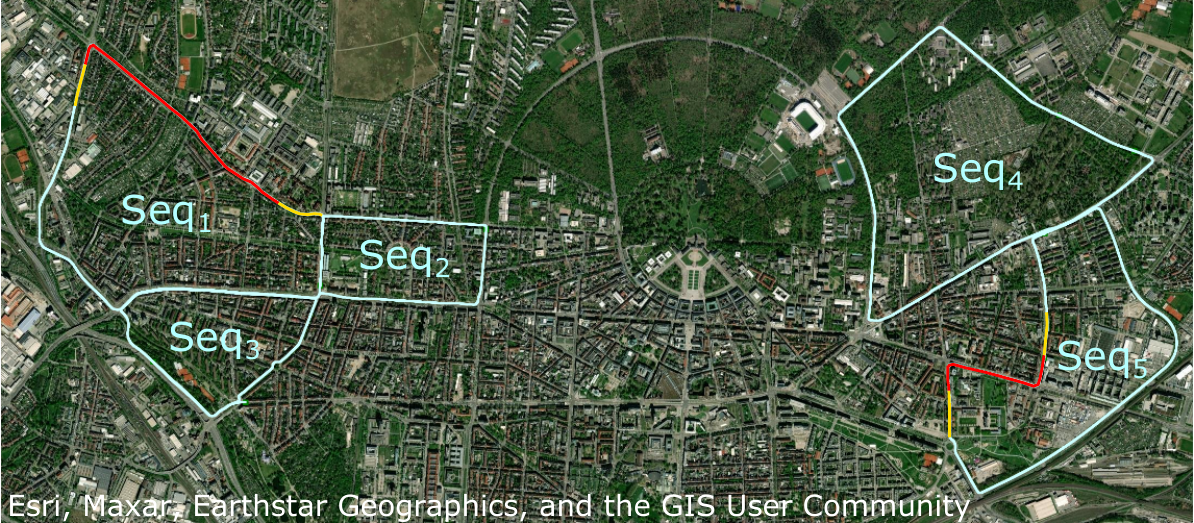}
    \caption{Spatial distribution of the sequences in Karlsruhe, Germany. The areas for the test set are depicted in \textcolor{red}{red}. The \textcolor{darkyellow}{yellow} areas show the geographic separation between training and test set.}
    \label{fig:sq_overview}
\end{figure}

\paragraph{Dataset Sequences and Split}

As shown in \cref{tab:DataRecording}, our dataset comprises five recorded sequences, spanning over 21 kilometers and containing more than 22,000 frames. The automatic parametric mapping approach described in \cref{sec:para_mapping} yields a total of 6,931 objects, each represented by the semantically tailored geometric primitives.

Recent research~\cite{Lilja2024CVPR} has demonstrated that many contemporary map perception studies rely on evaluation benchmarks that do not account for the geographic overlap between the train and test set, which significantly impacts evaluation results. Building upon this insight, we ensure a strict geographic separation between the training and testing sets, maintaining a minimum distance of 100 meters between them. To enhance the diversity of the test set, we select two distinct urban areas: a relatively rural region and a densely populated area near the city center. \cref{tab:DataRecording} provides an overview of the dataset split, while \cref{fig:sq_overview} visualizes its spatial distribution. Due to the high correlation between consecutive frames, we evaluate on every fifth frame in the test set to reduce redundancy.

\paragraph{Task Definition and Metrics}

We evaluate our cross-modal domain adaptation system with the three tasks of 2D semantic segmentation, 3D semantic segmentation and panoptic segmentation. This covers common computer vision task modalities and the common input representations of LiDAR data, range images and 3D point clouds. The evaluation regimes follow established definitions and metrics. For 2D semantic segmentation we report the IoU as calculated in \cite{Everingham2014ThePV} and for 3D semantic segmentation we follow the task definition and evaluation metric of \citet{Behley2019}. Panoptic segmentation is evaluated with the panoptic segmentation quality as defined in \cite{kirillov_panoptic_2019}.
All metrics are calculated with the three object classes of poles, traffic lights and traffic signs, dropping the background class which is present during training. For panoptic segmentation this results in reporting only the metrics for the "thing" classes, as the background is the only "stuff" class in the dataset.


\begin{table*}[t]
\centering
\caption{Quantitative results for the 2D semantic segmentation and panoptic segmentation tasks.}
\begin{tabular}
{@{\extracolsep{\fill}}llcrrrrrrr@{}}
\toprule
\multicolumn{2}{c}{\textbf{Experiments}} & \multirow{2}{*}{\shortstack[c]{\textbf{Motion} \\ \textbf{Comp.}}} & \multicolumn{4}{c}{\textbf{Semantic}} & \multicolumn{3}{c}{\textbf{Panoptic}} \\ 
\cmidrule(lr){4-7} \cmidrule(l){8-10}
&  &  & $\textrm{IoU}_{\textrm{Po}}$ & $\textrm{IoU}_{\textrm{TL}}$ & $\textrm{IoU}_{\textrm{TS}}$ & $\textbf{mIoU}$ & $\textrm{SQ}_{\textrm{th}}$ & $\textrm{RQ}_{\textrm{th}}$ & $\textbf{PQ}_{\textbf{th}}$\\
\cmidrule(r){1-3} \cmidrule(lr){4-7} \cmidrule(l){8-10}
\multirow{7}{*}{\shortstack[l]{\textbf{Baselines} \\ {[10 Hz, 50 m]}}} 
                                        & XD-B1             & \xmark & 9.6&9.1&9.2& 9.3& 65.2& 7.4& 4.8 \\
                                        & XD-B2             & \xmark & 15.0&18.6&17.0& 16.9 & 63.4& 11.7 & 7.4 \\
                                        & XD-MAP            & \xmark & 34.7&39.4&29.2& 34.4 & 67.8 & 35.9 & 24.4 \\
                                        \cmidrule(r){2-3} \cmidrule(lr){4-7} \cmidrule(l){8-10}  \vspace{-0.4cm}\\
                                        & XD-B1             & \cmark & 9.9 & 11.3 & 8.1 & 9.8 & 64.3 & 8.9 & 5.8\\
                                        & XD-B2             & \cmark & 17.9 & 18.6 & 16.1 & 17.5 & 63.6 & 12.6 & 8.0 \\
                                        & XD-MAP            & \cmark & \textbf{37.1} & \textbf{42.3} & \textbf{31.8} & \textbf{37.0} & \textbf{69.4}  & \textbf{39.6} & \textbf{27.5} \\

\cmidrule(r){1-3} \cmidrule(lr){4-7} \cmidrule(l){8-10} 
\multirow{7}{*}{\shortstack[l]{\textbf{Element Range} \\ {[XD-MAP, 10 Hz]}}}
                                        & 30 m              & \xmark & 35.4&42.9&30.2& 36.2 & 68.1 & 41.2 & 28.1 \\
                                        & 50 m              & \xmark & 34.7&39.4&29.2& 34.4 & 67.8 & 35.9 & 24.4 \\
                                        & 70 m              & \xmark & 32.5&37.1&29.1& 32.9 & 67.1 & 33.0 & 22.1 \\
                                        \cmidrule(r){2-3} \cmidrule(lr){4-7} \cmidrule(l){8-10} \vspace{-0.4cm}\\
                                        & 30 m              & \cmark & 40.0 & 45.4 & 31.9 & 39.1& 70.3 & 47.4& 33.4\\
                                        & 50 m              & \cmark & 37.1 & 42.3 & 31.8 & 37.0 & 69.4 & 39.6 & 27.5\\
                                        & 70 m              & \cmark & 36.4 & 40.1 & 30.8 & 35.7 & 69.0 & 35.7 & 24.7 \\

\cmidrule(r){1-3} \cmidrule(lr){4-7} \cmidrule(l){8-10}  
\multirow{3}{*}{\shortstack[l]{\textbf{Sensor Frequency} \\ {[XD-MAP, 50 m]}}} 
                                        & 0.5 Hz            & \cmark & 33.9 & 38.5 & 28.4 & 33.6 & 68.7 & 36.1 & 24.8 \\
                                        & 2 Hz              & \cmark & \textbf{37.4} & 41.2 & 30.8 & 36.4 & 69.2 & 39.1 & 27.1 \\
                                        & 10 Hz             & \cmark & 37.1 & \textbf{42.3} & \textbf{31.8} & \textbf{37.0} & \textbf{69.4}  & \textbf{39.6} & \textbf{27.5} \\
                                        
\bottomrule
\end{tabular}
\label{table_big}
\end{table*}

\begin{table}
\centering
\caption{Quantitative results for 3D semantic segmentation. For configuration parameters not specified, the default configuration is {[XD-MAP, 10 Hz, 50 m]} with motion compensation as in \cref{table_big}.}
\begin{tabular}
{@{\extracolsep{\fill}}lrrrr@{}}
\toprule
\textbf{Exp.} & \multicolumn{4}{c}{\textbf{Semantic 3D}} \\ 
\cmidrule(lr){2-5}
& $\textrm{IoU}_{\textrm{Po}}$ & $\textrm{IoU}_{\textrm{TL}}$ & $\textrm{IoU}_{\textrm{TS}}$ & $\textbf{mIoU}$ \\
\midrule
 XD-B1             & 0.5 & 5.1 & 17.1 & 7.6 \\
 XD-B2             & 0.5 & 7.6& 26.6 & 11.6 \\
 XD-MAP [MC:\xmark]    & 26.8 & 42.8& 30.4 & 33.3 \\
 XD-MAP            & \textbf{41.4} & \textbf{52.3} & \textbf{38.0} & \textbf{43.9} \\
 \midrule
 30 m              & 40.5 & 53.4 & 38.0 & 44.0 \\
 50 m              & 41.4 & 52.3 & 38.0 & 43.9 \\
 70 m              & 40.2 & 50.9 & 37.8 & 43.0 \\
 \midrule
 0.5 Hz            & 39.9& 50.3 & 34.1 & 41.5 \\
 2 Hz              & 41.6 & 51.9 & 37.1 & 43.5 \\
 XD-MAP            & \textbf{41.4} & \textbf{52.3} & \textbf{38.0} & \textbf{43.9} \\
                                        
\bottomrule
\end{tabular}
\label{table_3d_seg}
\end{table}

\begin{figure*}[h]
    \centering
    \renewcommand{\arraystretch}{1.1} 

    \begin{subfigure}[b]{0.99\textwidth}
        \centering
        \includegraphics[width=0.48\textwidth]{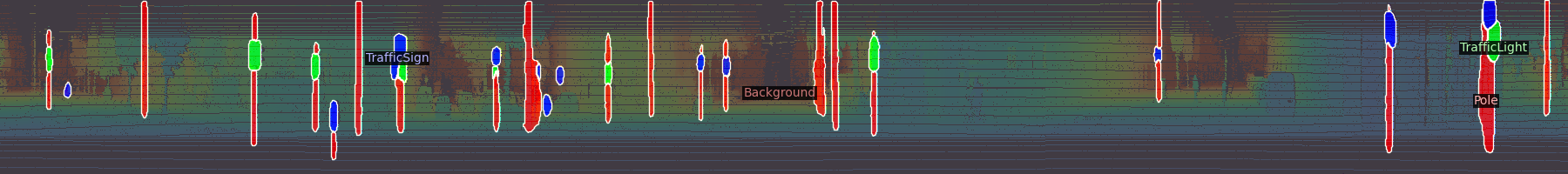}
        \includegraphics[width=0.48\textwidth]{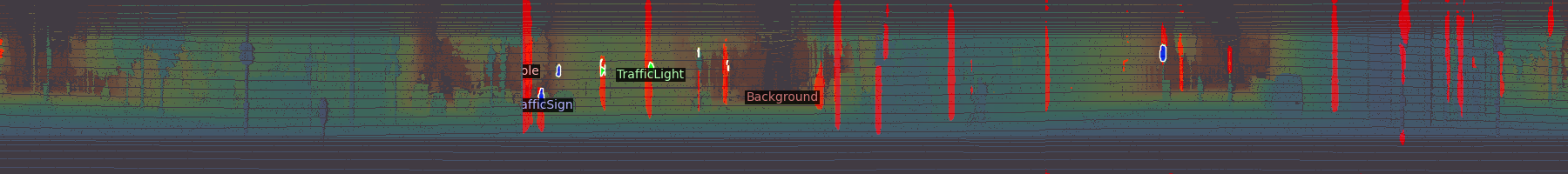}\\
        \includegraphics[width=0.48\textwidth]{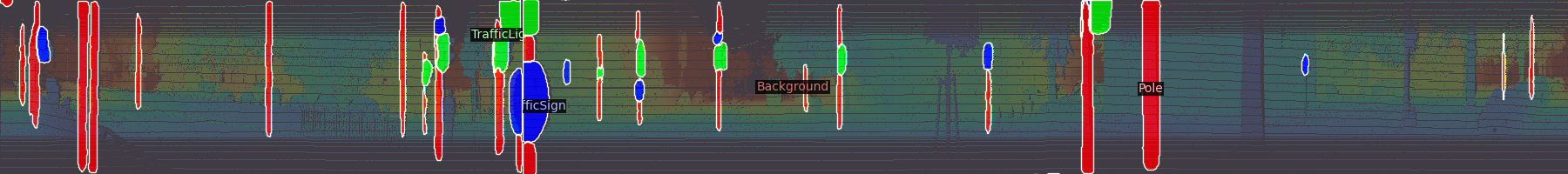}
        \includegraphics[width=0.48\textwidth]{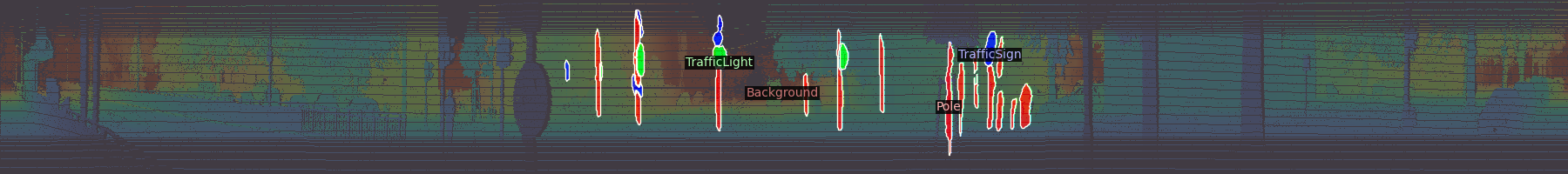}
        \caption{Semantic segmentation. Comparison of XD-MAP [10HZ, 50m] (left) and XD-B2 [10HZ, 50m] (right)}
    \end{subfigure}

    \begin{subfigure}[b]{0.99\textwidth}
        \centering
        \includegraphics[width=0.48\textwidth]{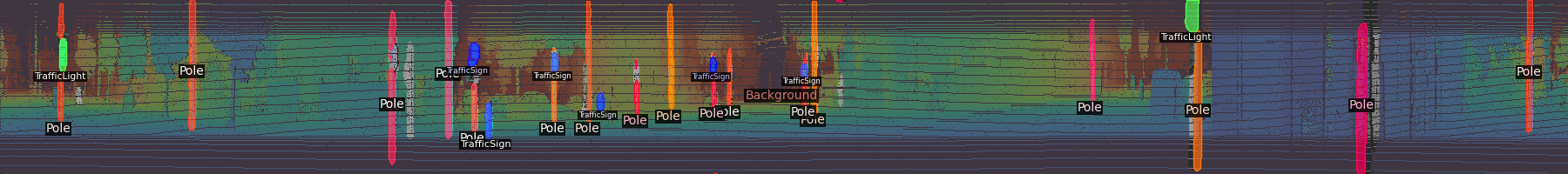}
        \includegraphics[width=0.48\textwidth]{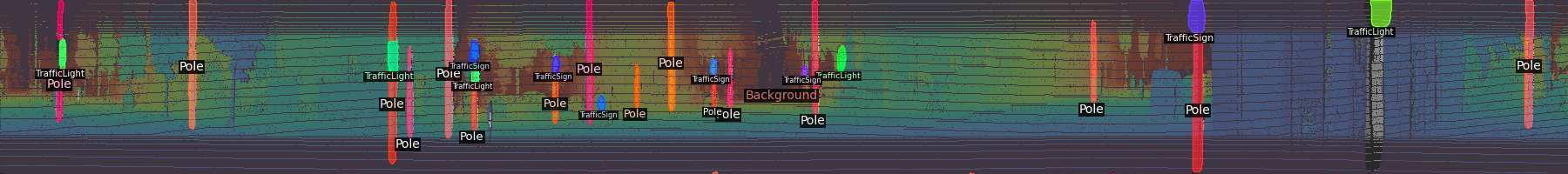}\\
        \includegraphics[width=0.48\textwidth]{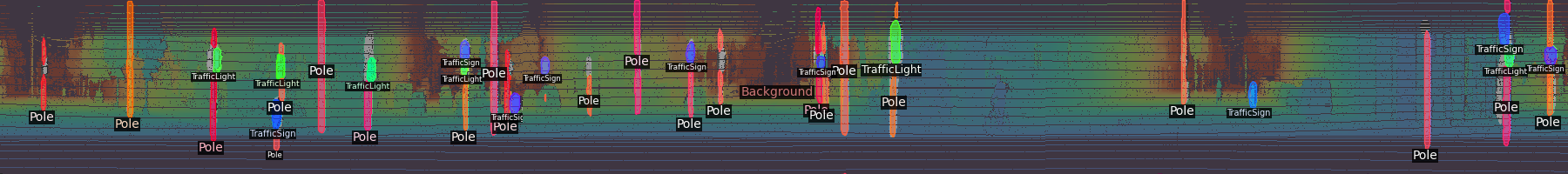}
        \includegraphics[width=0.48\textwidth]{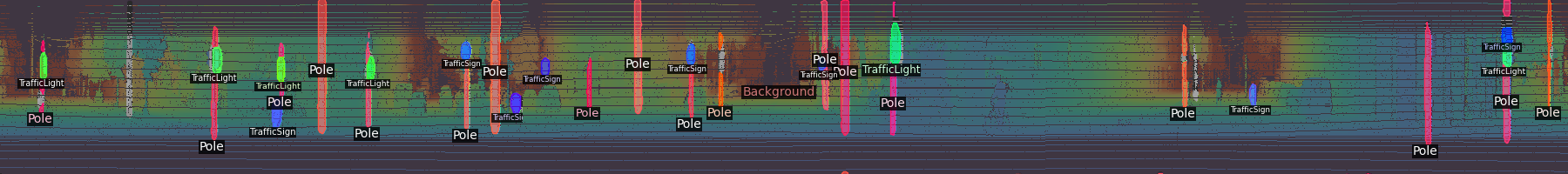}\\
        \includegraphics[width=0.48\textwidth]{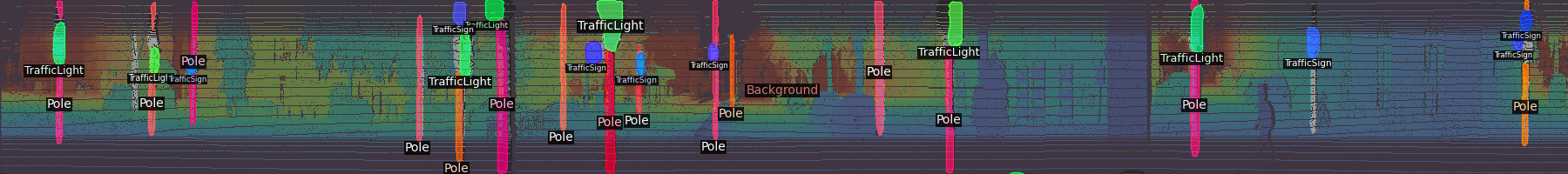}
        \includegraphics[width=0.48\textwidth]{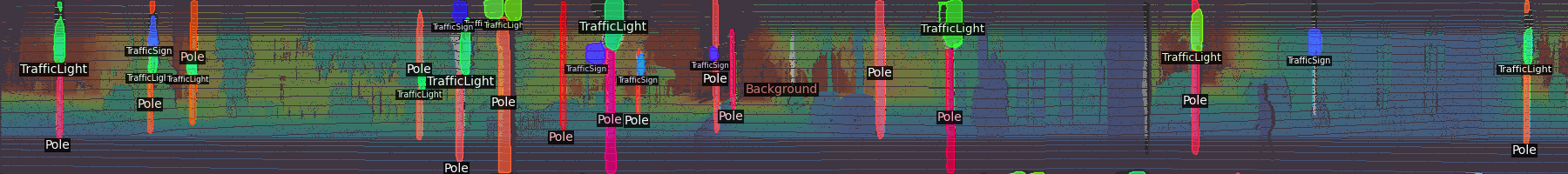}\\
        \caption{Panoptic segmentation. Comparison of XD-MAP [10HZ, 50m] (left) and XD-MAP [0.5HZ, 50m] (right)}
    \end{subfigure}

    \caption{Qualitative results of 2D perception.}
    \label{fig:results_2D}
\end{figure*}

\begin{figure*}[h]
    \centering
    \renewcommand{\arraystretch}{1.1} 
    \begin{tabular}{c c c c}  
        \textbf{GT XD-MAP} & \textbf{GT XD-B2} & \textbf{Prediction XD-MAP} & \textbf{Prediction XD-B2} \\ 
        
        \includegraphics[width=0.225\textwidth]{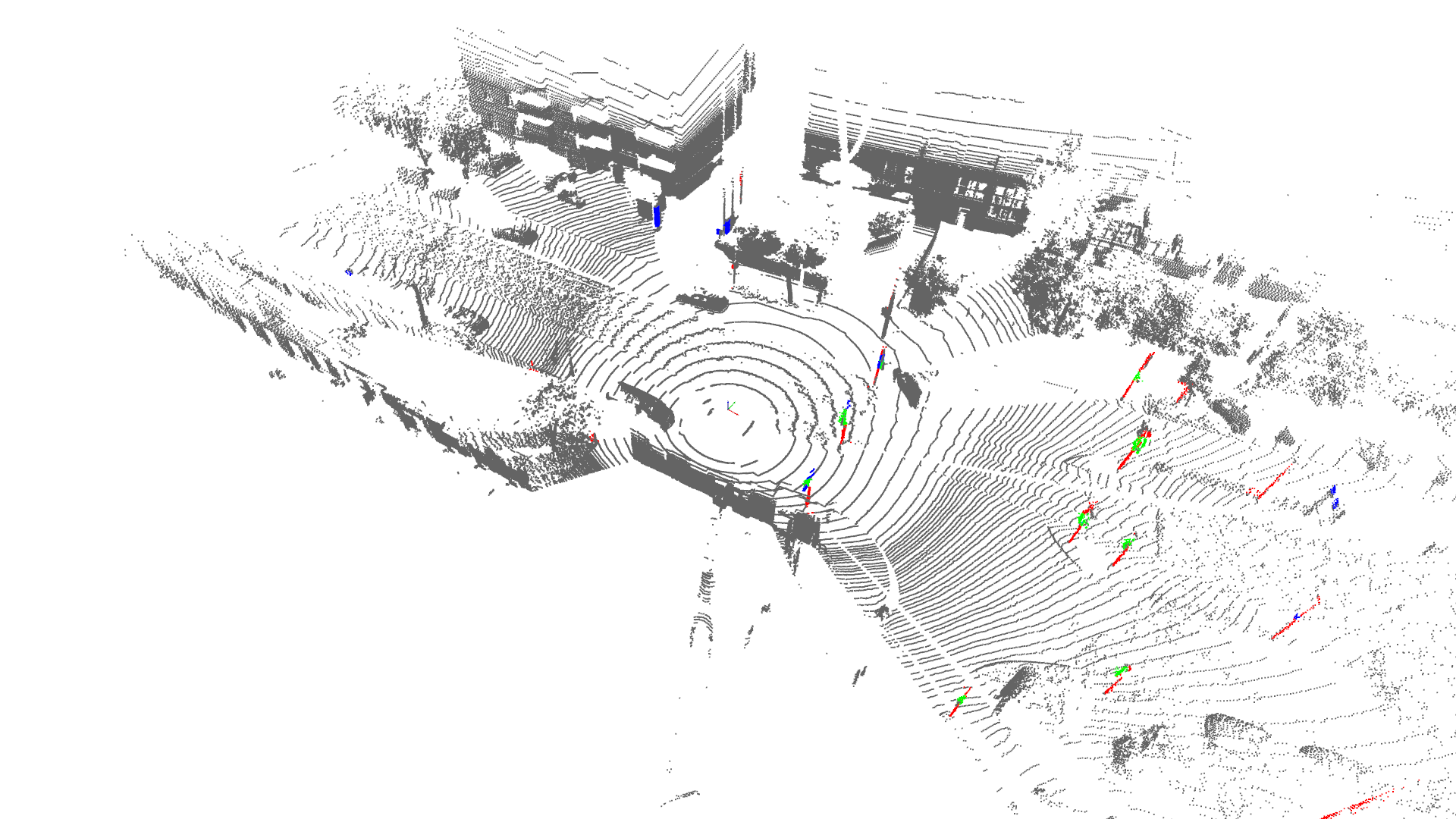} & 
        \includegraphics[width=0.225\textwidth]{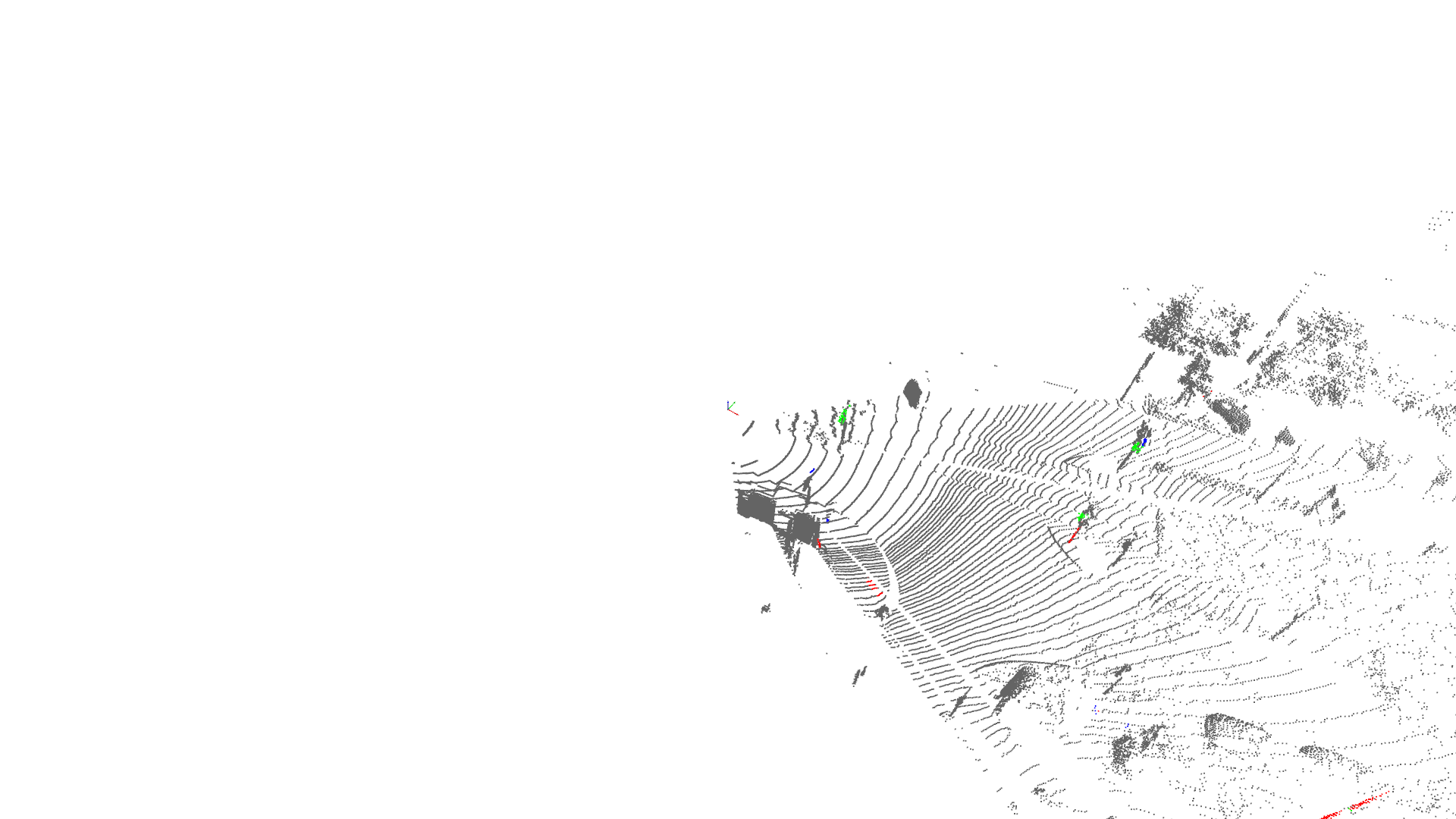} & 
        \includegraphics[width=0.225\textwidth]{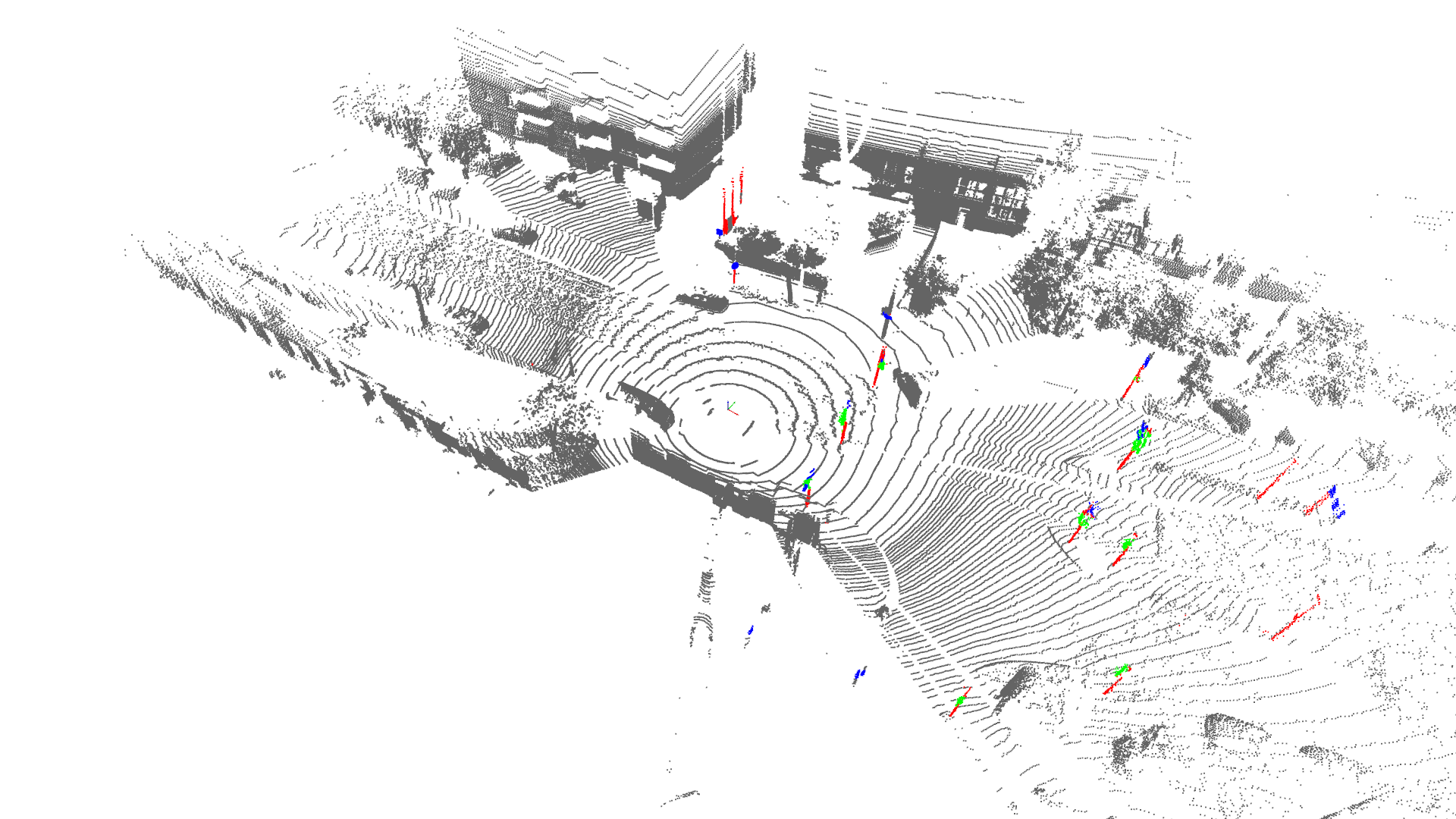} &  
        \includegraphics[width=0.225\textwidth]{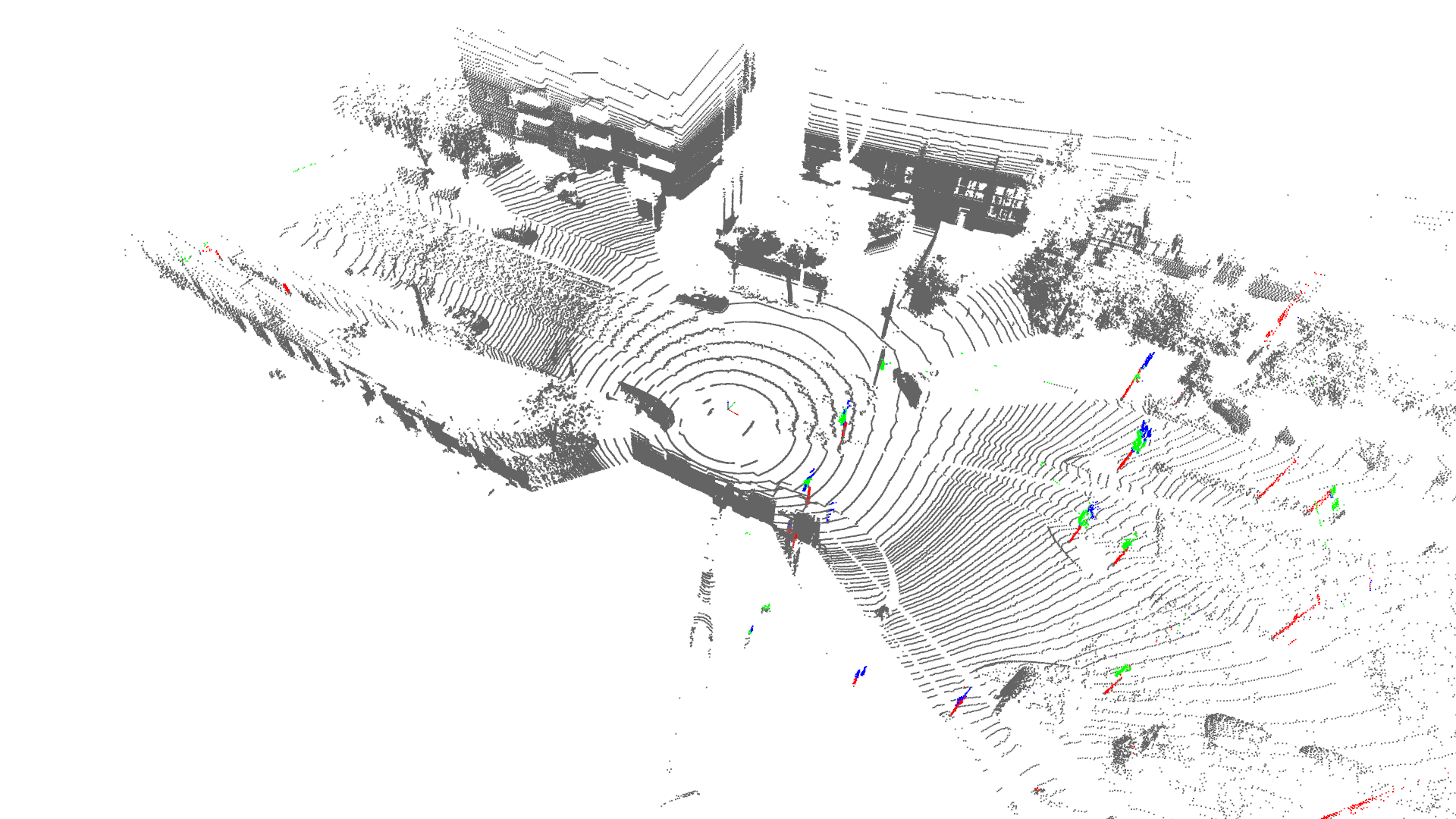} \\ 
        \includegraphics[width=0.225\textwidth]{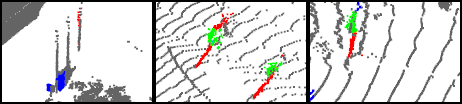} & 
        \includegraphics[width=0.225\textwidth]{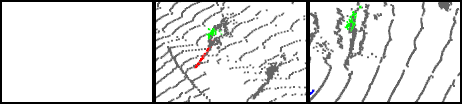} & 
        \includegraphics[width=0.225\textwidth]{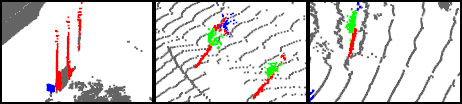} &  
        \includegraphics[width=0.225\textwidth]{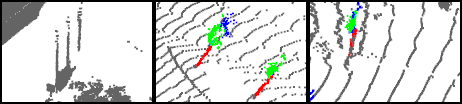} \\

        \includegraphics[width=0.225\textwidth]{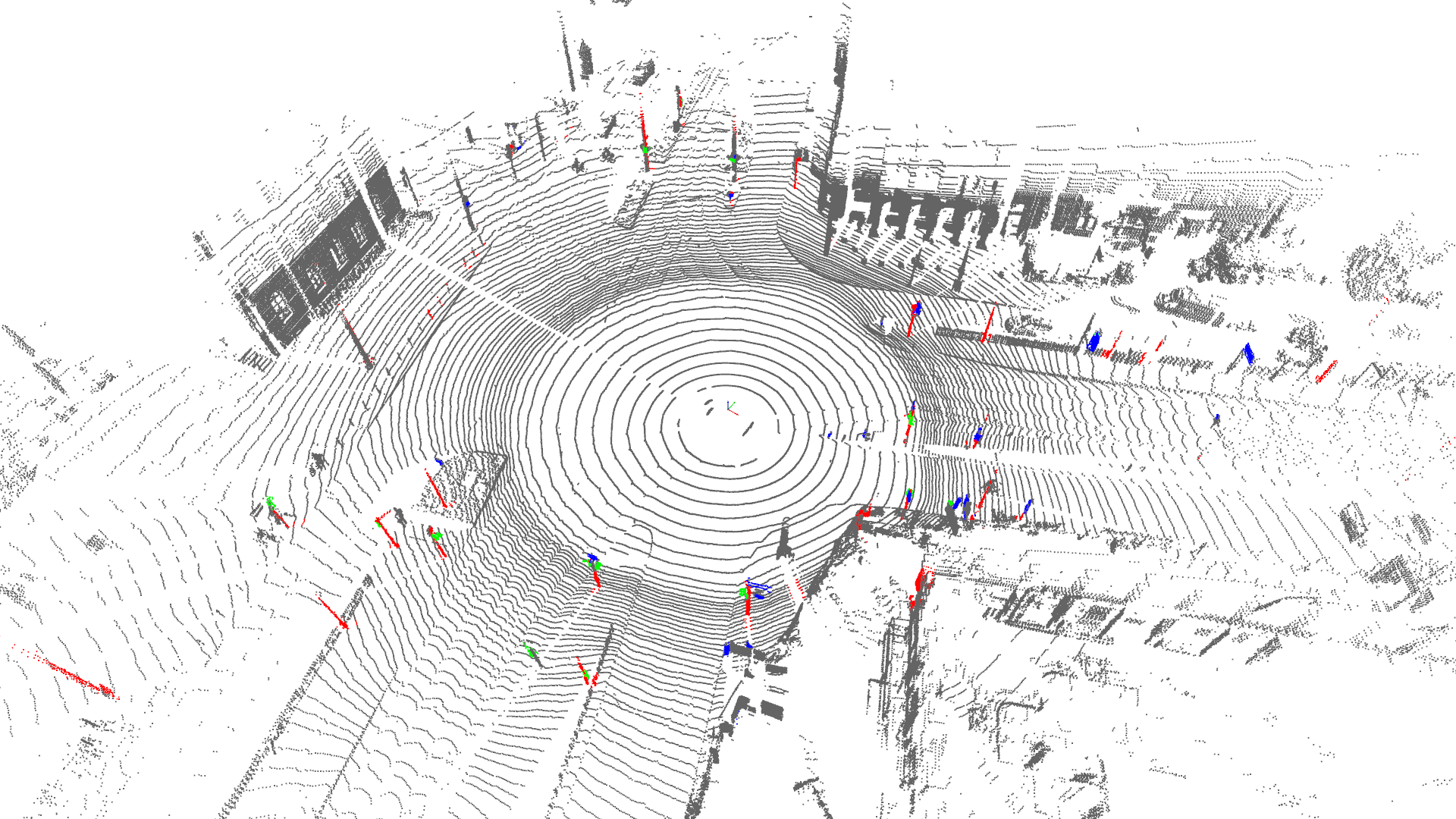} & 
        \includegraphics[width=0.225\textwidth]{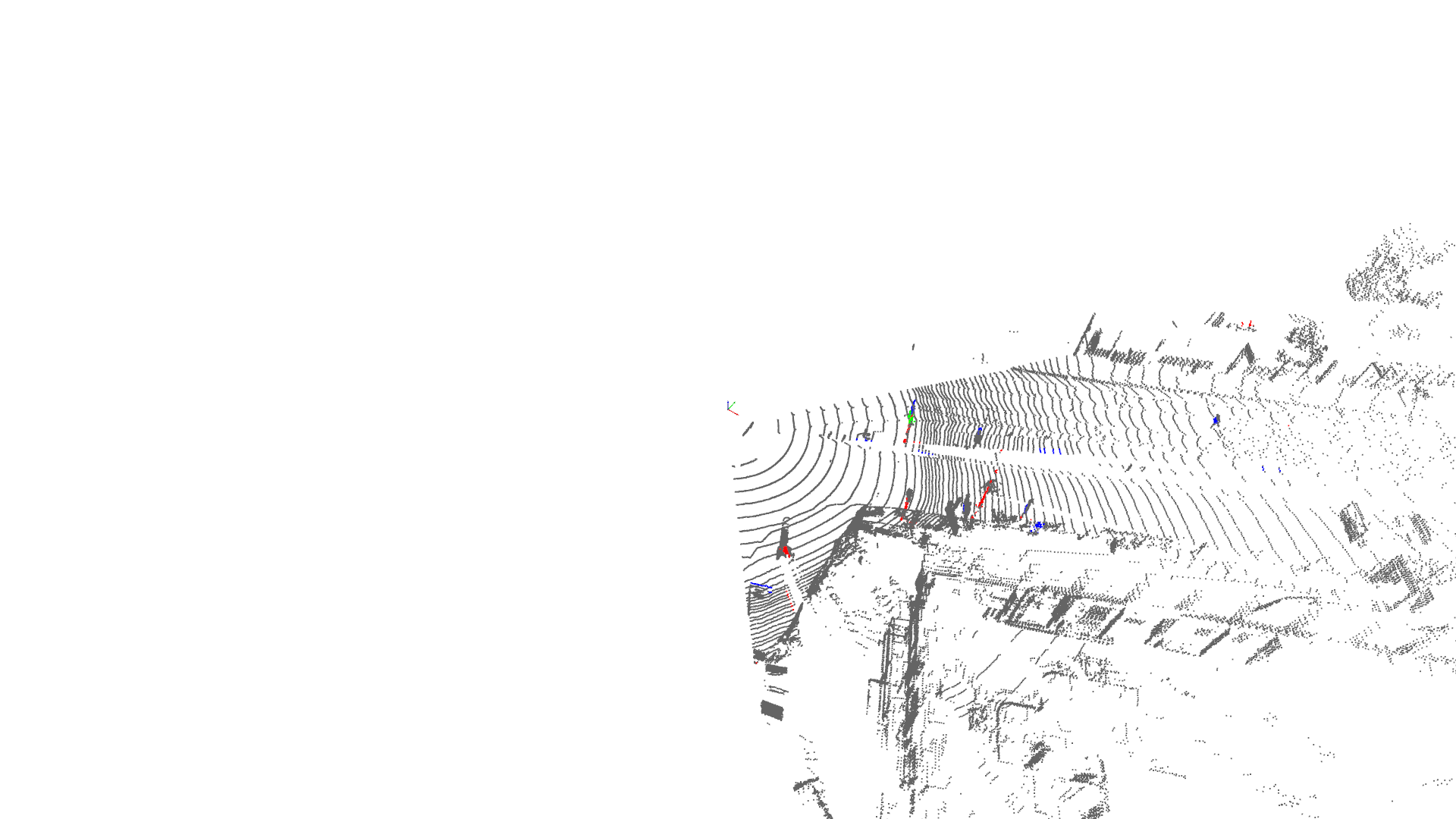} & 
        \includegraphics[width=0.225\textwidth]{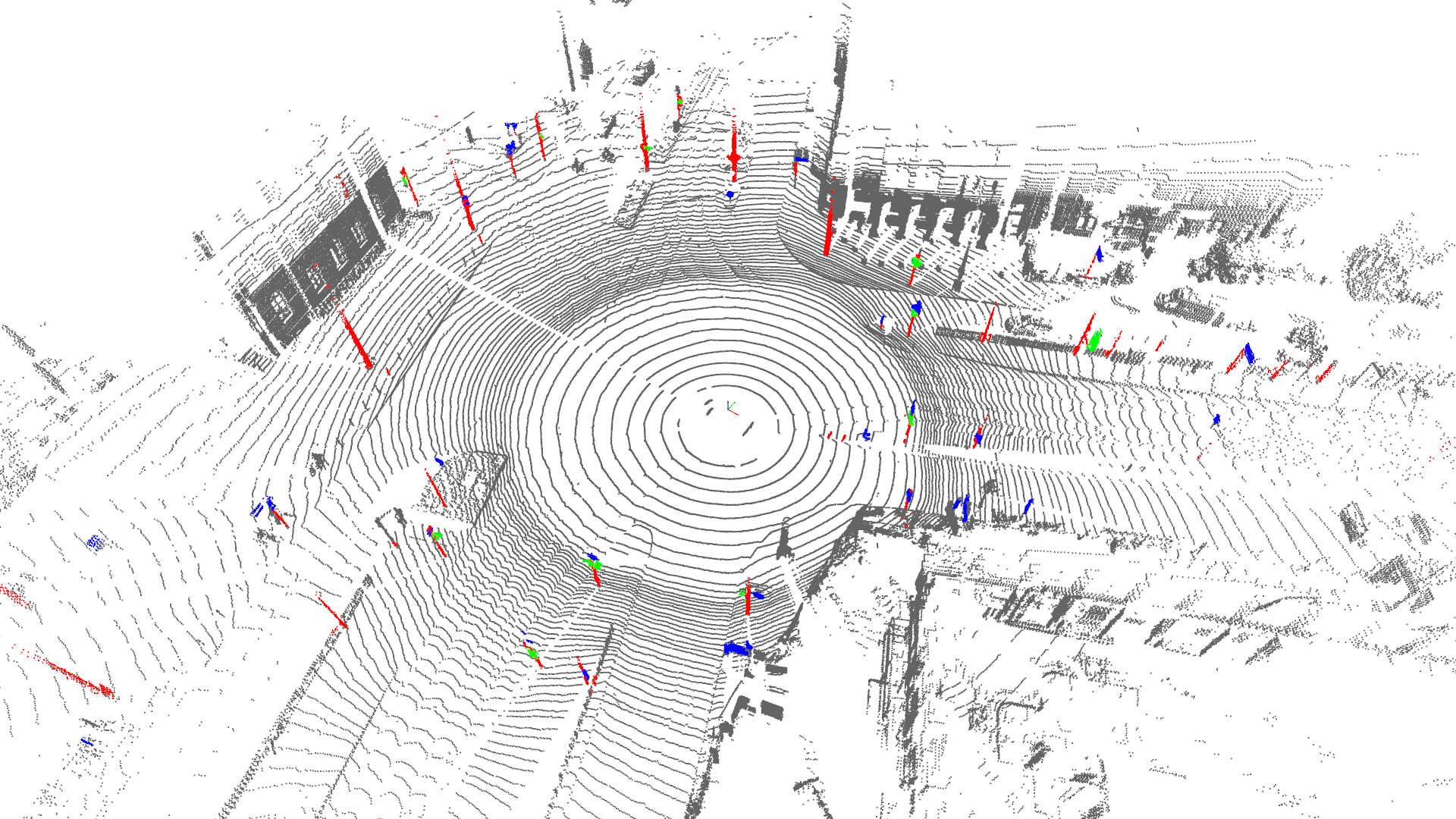} &  
        \includegraphics[width=0.225\textwidth]{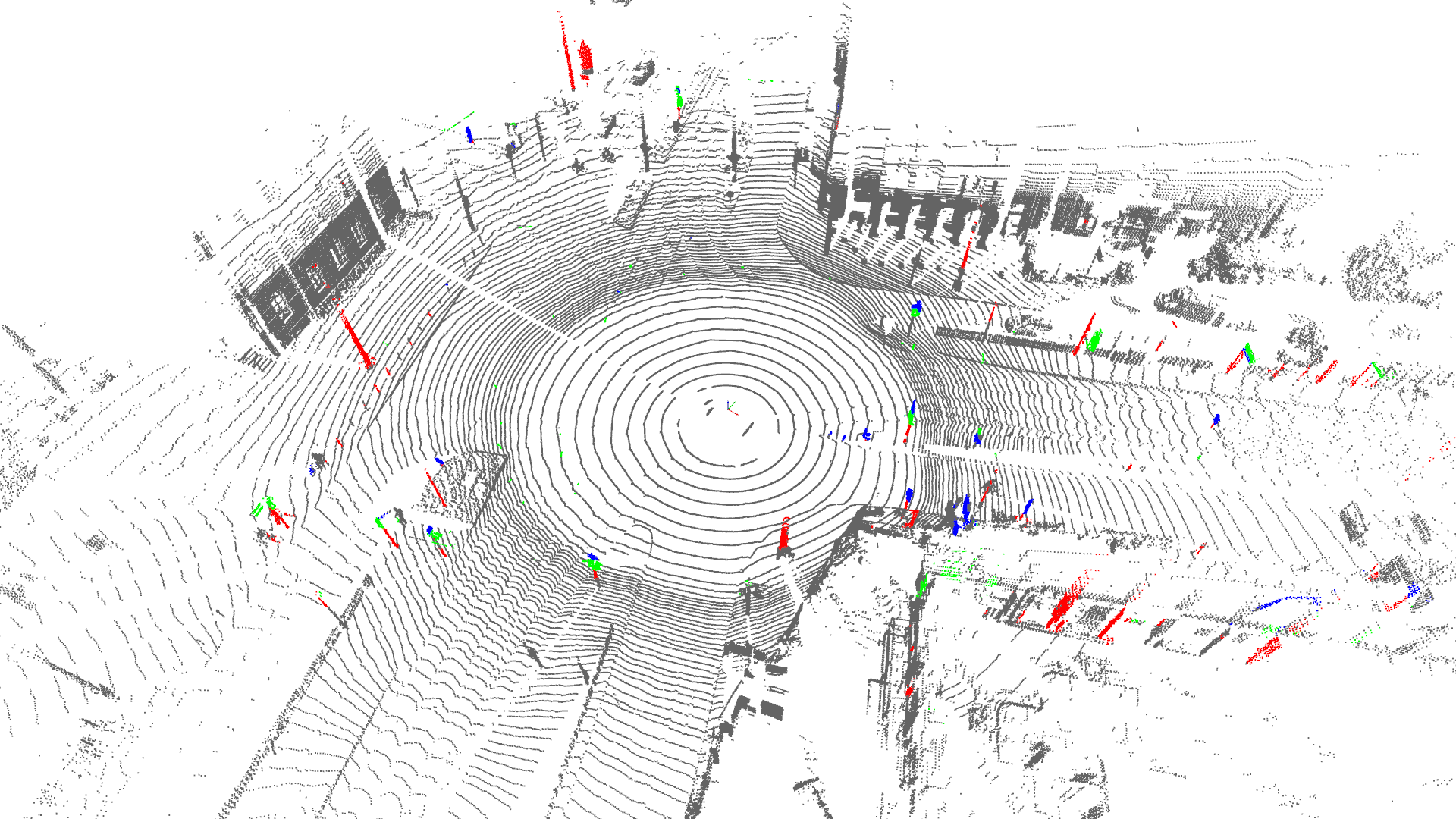} \\ 

        \includegraphics[width=0.225\textwidth]{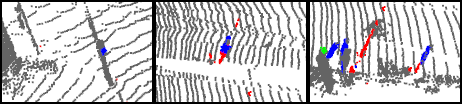} & 
        \includegraphics[width=0.225\textwidth]{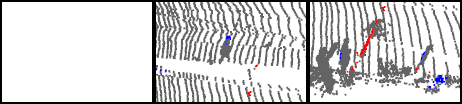} & 
        \includegraphics[width=0.225\textwidth]{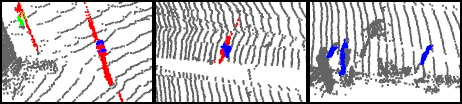} &  
        \includegraphics[width=0.225\textwidth]{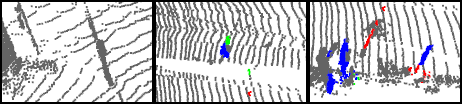} \\ 

    \end{tabular}
    
    \caption{Qualitative results of 3D segmentation. Comparison of pseudo labels and predictions of XD-Map and XD-B2.}
    \label{fig:3x5_images}
\end{figure*}

\subsection{Models and Hyperparameter Setting}

\paragraph{Mask2Former}

We select Mask2Former \cite{mask2former} as the model architecture for 2D semantic segmentation and panoptic segmentation. Most hyperparameters are based on the model config for COCO \cite{coco2014}, except for some changes related to the differences from range images to RGB images. To account for these differences we disable RGB augmentations and increase the learning rate multiplier of the pretrained backbone to 0.4 from 0.1. The model is trained with a base learning rate of 0.0001, using 80,000 iterations with a batch size of 10 for semantic segmentation and 68,000 iterations with a batch size of 16 for panoptic segmentation.

\paragraph{Cylinder3D}
For the task of 3D segmentation, we chose the Cylinder3D~\cite{zhu2021cylindrical} architecture. We adopt the hyperparameters for training on SemanticKITTI~\cite{thisanke2023semantic}, except for the number of epochs. The model is trained for 12 epochs with a batch size of 16. AdamW optimizer is used with a base learning rate of 0.001 and weight decay of 0.01. 

\subsection{Experiment Layout}

In our experiments, we first show XD-MAP's performance \wrt two baselines, underlining the effectiveness of the proposed approach.
Additionally, we present ablation studies for the three parameters with the greatest practical importance in the autonomous driving domain: motion compensation, map element range and sampling frequency.

\paragraph{Motion Compensation}

Due to the ego-motion, LiDAR points are distorted and cannot be trivially referenced to a single pose.
We mitigate this 
by inverting the sensor motion \wrt the reference pose.
Assuming static objects, this shifts all LiDAR points as if they were measured from the reference pose.
Since our labels are not pixels in an individual image, but generated from mapped geometric primitives in 3D space, motion compensation impacts the congruence of the labels with the range image. 
We examine its influence comparing trainings with and without motion compensation.

\paragraph{Range of Map Elements}

Because the range of detected map elements is important for downstream applications, we investigate the performance of XD-MAP for three ranges of \SI{30}{m}, \SI{50}{m}, and \SI{70}{m}.
Since motion compensation might affect farther points more severely, we examine the interplay of motion compensation and range.

\paragraph{Sampling Frequency}

Sensor or sampling frequency directly correlates with dataset size, causing a strong motivation to keep the sampling frequency as small as possible.
While a higher sampling frequency enhances real-world augmentation, it also introduces a higher degree of correlation between consecutive data samples. 
To investigate the influence of sampling frequency on model performance, we compare rates of 0.5 Hz, 2 Hz, and 10 Hz.

\subsection{Results}

\cref{table_big} contains the results of the 2D semantic segmentation and panoptic segmentation task for XD-MAP, the baselines and the investigated element ranges and sampling frequencies. Furthermore, \cref{fig:results_2D} and \cref{fig:3x5_images} show qualitative results.

\paragraph{Comparison with Baselines}

XD-MAP shows strong performance increases of +27.2 and +19.5 mIoU points in the 2D semantic segmentation task compared to the single shot baselines XD-B1 and XD-B2, respectively. 
Similar increases are consistent across all element classes. 
Metrics for panoptic segmentation mirror this increase, with an improvement of +21.7 and +19.5 $\textrm{PQ}_{\textrm{th}}$ compared to XD-B1 and XD-B2.
Looking at the other metrics, it can be seen that this gain can almost entirely be attributed to the gain in recognition quality $\textrm{PQ}_{\textrm{th}}$, meaning models trained with XD-MAP have much higher precision and recall of detected elements.
These results clearly demonstrate the benefits of a systematic approach of a highly accurate SLAM in combination with parametric mapping of geometric primitives for cross-modal domain adaptation in contrast with simpler approaches limited by camera field of view or inaccuracies of the pseudo-labels.

\paragraph{Effect of Map Element Range}

The performance of the network, as expected, decreases with increasing element range of the labels, both for the 2D semantic segmentation and panoptic segmentation.
For the semantic segmentation, the difference between mIoU of \SI{30}{m} range and \SI{70}{m} amounts to -3.4, while for the panoptic segmentation it is a bit larger at -8.7 $\textrm{PQ}_{\textrm{th}}$. 
Interestingly though, this decrease mainly stems from the decrease in recognition quality $\textrm{RQ}_{\textrm{th}}$, while the segmentation quality $\textrm{SQ}_{\textrm{th}}$ decreases only slightly.
This suggests that the model has trouble with detecting the small elements appearing with increasing range.

\paragraph{Effect of Sampling Frequency}

A similar expected trend of decreasing performance can also be observed with decreasing sensor frequency.
However, for a frequency of 2 Hz the decrease only amounts to -1.0 mIoU for semantic segmentation and -0.4 $\textrm{PQ}_{\textrm{th}}$ for 
panoptic segmentation.
This implies that for use cases with stricter requirements on dataset size, similar performance can be reached with only a quarter of needed storage.

\paragraph{Effect of Motion Compensation}

Disabled motion compensation causes a consistent decrease in performance across tasks, ranging from -0.5 to -2.9 mIoU in the case of 2D semantic segmentation.
As the model does not have the ego velocity available, which is required to properly apply motion compensation, it is unable to compensate the additional noise by itself. 
Even though it would be theoretically possible for a model to learn and correct the additional noise pattern introduced my motion compensation, in practice the model still became disturbed.
These results further underline the importance of careful dataset curation, including the reduction of sensor noise when possible.

\paragraph{3D Semantic Segmentation}
The results for 3D semantic segmentation follow the trends of the other tasks but reveal important differences.
XD-MAP displays large improvements in mIoU from the single-shot baselines which both, yet XD-B1 in particular, fails to label many objects, leading models to predict mostly background.
While increasing the element range reduces performance, the drop is smaller than in 2D, suggesting that long-range segmentation benefits from the richer 3D representation of point clouds.
In contrast, the absence of motion compensation leads to a much stronger performance decline, which could indicate that our 3D labeling is more dependent on a precise alignment of LiDAR points and HD map to ensure points are located in an object's frustum. This highlights a relative advantage of 2D labeling to be less sensitive to localization, mapping or calibration errors.
Finally, to the human eye, the prediction appears to surpass the ground truth annotation in many cases in the qualitative evaluation.

\section{Limitations}
While we have demonstrated XD-MAP for domain adaptation from camera to LiDAR, we argue that it is applicable for many sensor combinations, environments, and semantic classes. This comes with a few limitations: While we do not need an overlapping field of view of the sensors, it is required that both sensors perceive the same objects over time. Also the resolution of both sensors needs to be adequate to the size, level of detail and maximum distance of mapped objects. 
%
Since the baselines only provide a small field of view and visual inspection shows the much better accuracy of XD-MAP annotations, these were selected as our ground truth.
The XD-MAP pipeline also requires a highly accurate SLAM and multimodal calibration. Sensor suites with very noisy sensors or without accurate calibration may lack the prerequisites for semantic parametric mapping of instances.
Furthermore, objects that are not representable with geometric primitives are not able to be processed with the proposed system, which may limit its applicability outside of structured domains like roads in autonomous driving. 

 \section{Conclusion}

 \label{sec:conclusion}
We presented XD-MAP, a novel domain adaptation technique that makes it possible to transfer knowledge across sensing modalities without requiring similarity of sensor characteristics.
Neither is XD-MAP limited by the sensors' fields of view.
By fully automatically creating a highly accurate HD map whose parametric representation is tailored to the respective semantic class of objects, we can accumulate knowledge available in the source domain.
The geometric representations, cylinders for poles and traffic lights as well as shaped planes for road signs, enable generating pixel- and point-accurate pseudo labels for the target sensor.

We prove the effectiveness of our approach by transferring knowledge from panoptically labeled RGB images from a front-view camera to a \SI{360}{\degree} LiDAR as target.
Evaluating three complementary tasks, XD-MAP outperforms the best baseline by +19.5 mIoU for 2D semantic segmentation, +19.5 $\textrm{PQ}_\textrm{th}$ for 2D panoptic segmentation, and +32.3 mIoU in 3D semantic segmentation.
This shows that the method generalizes well across a comprehensive array of typical tasks.
For applications, XD-MAP hence makes the knowledge, which is captured in the broad variety of semantically rich and well-generalizing image datasets, such as Mapillary Vistas, available to still underrepresented or specialized sensors, such as LiDAR.
While we demonstrated XD-MAP with three exemplary classes of static objects, future work will extend XD-MAP to more semantic classes, including dynamic objects that require 4D reconstruction, as well as more sensors, such as imaging radar.

\section*{Acknowledgment}
The research leading to these results was partially funded by the German Federal Ministry for Economic Affairs and Energy (BMWE) and the European Union within the \emph{just better DATA \mbox{(jbDATA)}} project, grant number 19A23003H.
The authors gratefully acknowledge the computing time provided on the high-performance computer \emph{HoreKa} by the National High-Performance Computing Center at KIT (NHR@KIT). This center is jointly supported by the Federal Ministry of Education and Research and the Ministry of Science, Research and the Arts of Baden-Württemberg. HoreKa is partly funded by the German Research Foundation (DFG).

{
    \small
    \bibliographystyle{ieeenat_fullname}
    \bibliography{main}
}

\end{document}